# Forecasting Weakly Correlated Time Series in Tasks of Electronic Commerce

Lyudmyla Kirichenko
Applied Mathematics Department
Kharkiv National University of Radioelectronics
Kharkiv, Ukraine
lyudmyla.kirichenko@nure.ua

Tamara Radivilova
Infocommunication Engineering Department
Kharkiv National University of Radioelectronics
Kharkiv, Ukraine
tamara.radivilova@gmail.com

Illya Zinkevich
Applied Mathematics Department
Kharkiv National University of Radioelectronics
Kharkiv, Ukraine
illya.zinkevich@nure.ua

*Abstract — Forecasting of weakly correlated time series of conversion rate by methods of exponential smoothing, neural network and decision tree on the example of conversion percent series for an electronic store is considered in the paper. The advantages and disadvantages of each method are considered.*

*Keywords — Time series, conversion rate, forecasting, exponential smoothing, decision tree, long-term memory network*

## I. INTRODUCTION

Time series describe a wide range of phenomena, for example, they are the stock prices, solar activity, the overall incidence rate and much more. Economic indicators can also be considered as time series and you can try to find not visible at first glance laws, hidden periodicity, to predict the moments when peaks appear, etc. At the moment is urgent time-series analysis in the field of e-commerce. E-commerce is in process of development, which is facilitated by new technologies, services and tactical tools [1]. For successful sale in online stores, web analytics is used, which allows to work on optimization, increase conversion and attendance of the electronic store.

To "survive" and stand out among the many online stores, it is important to understand the user's behavior from the moment of the first arrival on the site: to track his movements, to know what products he looked at, put in the basket, where he clicked, what saw, the time he left, how and when he returned. Web-analytics will help in this, which involves ongoing collection, analysis and interpretation of data about visitors, work with basic metrics. Careful analysis of the online store and user behavior is a necessary stage of business development.

Quality web analytics of online store always begins with the visitor's way that he passed before making a purchase. The order processing consists of the following steps: 1) product search; 2) add item to shopping cart; 3) go to the checkout page; 4) fill out and submitting the form; 5) go to the page of the order, payment. The main task of analytics is to periodically find and fix the weak points in this chain. At each stage, the user can stop without having made a purchase. Each of these stages is represented in the form of a time series.

Conversion is the most important parameter that characterizes the effectiveness of the website promotion process. Conversion is the ratio of the number of users who made purchase of a product or service on your site to the number of users who came to your site for an advertising link, ad, or banner. For example, if a site was visited by 100 people, but only 2 people bought the product, then the conversion rate is equal to 2%. The increase of conversion percent depends on many factors: from the design of the page to its functionality. Monitoring conversion allows to understand in time that the e-shop needs to be improved. Conversion is the main metric in the web analytics of all commercial sites [2].

Analysis and forecasting of time series of daily value conversion percent plays crucial value for optimizing the efficiency of the online business. However, it should be noted that almost all of the classical methods of time series forecasting based on the calculation of the correlation between the time series values [3]. In the case of weakly correlated time series, and also in the case when the time series has sparse zero values structure, which is typical for many electronic sales sites, these methods do not fit or have a large error.

Neural network approach has been widely used to solve forecasting problems. Neural networks allow you to model complex relationships between data as result of learning by examples. However, the prediction of time series using neural networks has its drawbacks. First, for training the majority of neural networks, time series of a large length are required. Secondly, the result essentially depends on the choice of the architecture of the network, as well as the input and output data. Third, neural networks require preliminary data preparation, or preprocessing. Preprocessing is one of the key elements of forecasting: the quality of the forecast of a neural network can depend crucially on the form in which information is presented for its learning. The general goal of preprocessing is to increase the information content of inputs and outputs. An overview of the methods for selecting input variables and preprocessing is contained in [ 4].

Recently, for the analysis of the regularities of the time series, the methods of Data Mining and machine learning [5] have been increasingly used to detect various patterns in the time series. In this case, logical methods are of particular value in the detection of such patterns. These methods allow us to find logical if-then rules. They are suitable for analyzing and predicting both numerical and symbolic sequences, and their results have a transparent interpretation.

The goal of the presented work is to carry out a comparative analysis of weakly correlated time series forecasting, based on classical prediction methods and machine learning ones, using the data of a real online store.

## II. INPUT DATA

Input data in the work were daily data from the online sales site, which included the number of clicks on the site from social networks, the number of sales and the corresponding conversion rate. In addition, there was information about which language the customer used, from which country the order was and other data.

Fig. 1 (at the top) shows typical time series of conversion rate. Series of conversion rate are characterized by zero values, which significantly complicates forecasting the next day. The correlation function of the rate series is shown on Fig. 1 at the bottom. Obviously, there is no correlation between the time series values.

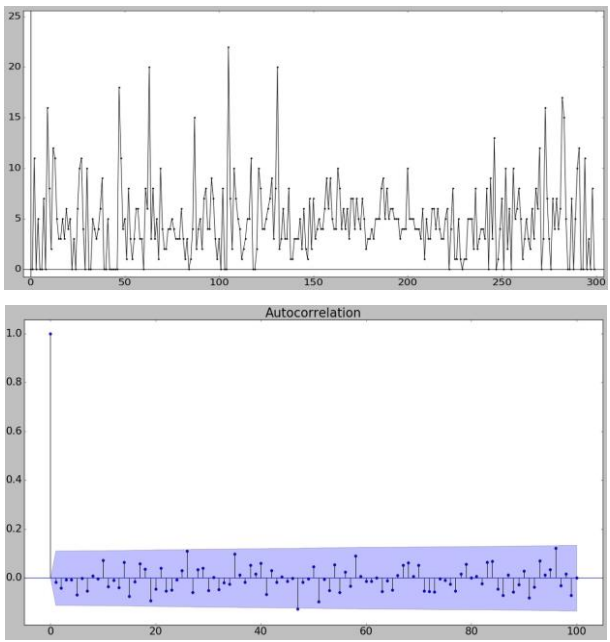

Fig. 1. Time series of conversion rate and the correlation function

Fig. 2 shows the histogram of the distribution density of typical conversion rate series. It is easy to see that the percent from 0 to 2 is the highest, then there is a more even distribution, but for each series of conversion rate there are bursts that are most difficult to forecast.

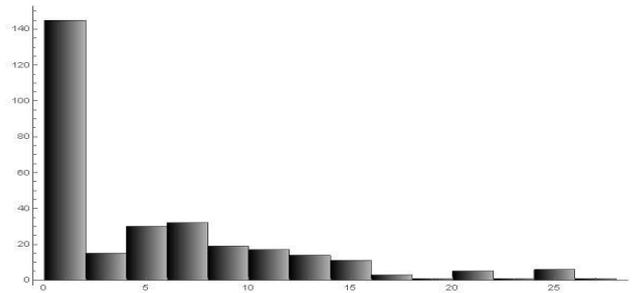

Fig. 2. Distribution density of typical conversion rate series

## III. FORECASTING METHODS

E-commerce is constantly evolving, it is facilitated by new technologies, services and tactical tools. Suppliers, range of buyers, assortment of goods change regularly, that leads to a rapid obsolescence of information. Therefore, forecasting methods that require time series of great length, such as, for example, autoregressive and moving average models, work poorly.

**Methods of exponential smoothing**. The basis for exponential smoothing is the idea of a constant revision of the forecast values as the actual ones arrive. The model of exponential smoothing assigns exponentially decreasing weights to observations as they become outdated [3]. Thus, the latest available observations have greater influence on the forecast value than older observations.

The model of exponential smoothing has the form:

$$Z(t) = S(t) + \varepsilon_t,$$
$$S(t) = \alpha \cdot Z(t-1) + (1-\alpha) \cdot S(t-1),$$

where $\alpha$ is smoothing factor; $0 < \alpha < 1$; $Z(t)$ is projected time series; $S(t)$ is smoothed time series; initial conditions are defined as $S(1) = Z(0)$. In this model, each subsequent smoothed value $S(t)$ is the weighted average between the previous value of the time series $Z(t-1)$ and the previous smoothed value $S(t-1)$.

**The methods of data maning** are an extremely broad and dynamically developing field of research using a huge number of theoretical and practical methods. One of the methods of machine learning is the decision tree method. The decision tree is a decision support tool used in statistics and data analysis for predictive models. The structure of the tree is "leaves" and "branches". On the edges ("branches") of the decision tree, attributes are recorded, on which the objective function depends, in the "leaves" the values of the objective function are recorded, and the remaining nodes contain attributes for which cases are distinguished.

To classify a new case, it is necessary to go down the tree to the leaves and give the appropriate value. Similar decision trees are widely used in intellectual data analysis. The goal is to create a model that predicts the value of the objective variable based on several input variables.

Each leaf represents the value of the objective variable which was modified during the movement from the root to the leaf. Each internal node corresponds to one of the input variables. The tree can also be "learned" by dividing the original sets of variables into subsets based on testing attribute values. This is process that is repeated on each of the received subsets.

The recursion terminates if the subset in the node has the same values of the objective variable, thus it does not add value for predictions. In intellectual data analysis, decision trees can be used as mathematical and computational methods to help describe, classify and summarize a set of data that can be written as follows: $(x, Y) = (x_1, x_2, x_3, \ldots, x_k, Y)$. The dependent variable $Y$ is the objective variable that needs to be analyzed, classified and generalized. A vector $x$ consists of input variables $x_1, x_2, x_3$, etc., which are used to perform this task [5,7].

**Network of long shot term memory** (LSTM). This is a special kind of recurrent neural networks, which are capable of learning long-term dependencies. LSTM are specifically designed to avoid the problem of long-term dependencies. Remembering information for a long period of time is practically their default behavior, not attempts to do. All recurrent neural networks have the form of a repeating module circuit of a neural network. In standard recurrent neural networks these repeating modules will have a very simple structure. LSTMs also have this chain like structure, but the repeating module has a different structure. Instead of having a single neural network layer, there are four, interacting in a very special way. Currently LSTM work incredibly well on a wide variety of tasks and are widely used [4, 6].

**The forecast errors**. To obtain quantitative characteristics of the comparative analysis of the models, the following characteristics of forecast errors were chosen [3]. The Mean Absolute Deviation (MAD) measures the accuracy of the forecast by averaging the values of the forecast errors. Using MAD is most useful when the analyst needs to measure the forecast error in the same units as the original series. This error is calculated as follows:

$$MAD = \frac{1}{n}\sum_{t=1}^{n}|X(t) - \hat{X}(t)|.$$

The average deviation (Mean Deviation, MD) allows you to see how the forecast value is overvalued or undervalued on average:

$$MD = \frac{1}{n}\sum_{t=1}^{n}X(t) - \hat{X}(t)$$

Mean squared error (MSE) is another way of estimating the forecasting method. Since each deviation value is squared, this method emphasizes large forecast errors. The MSE error is calculated as follows:

$$MSE = \frac{1}{n}\sum_{t=1}^{n}(X(t) - \hat{X}(t))^2.$$

The Mean Absolute Percentage Error (MAPE) is calculated by finding the absolute error at each time and dividing it by the actual observed value, with subsequent averaging of the obtained absolute percent errors. This approach is useful when the size or value of the predicted value is important for estimating the accuracy of the forecast. MAPE emphasizes how large the forecast errors are in comparison with the actual values of the series. This error is calculated as follows:

$$MAPE = \frac{1}{n}\sum_{t=1}^{n}\frac{|X(t) - \hat{X}(t)|}{X(t)}.$$

IV. RESEARCH RESULT

To build models and a neural network, the Python language with libraries implementing machine learning methods was used. Python is a general purpose programming language, which means that people have built modules to create websites, interact with a variety of databases, and manage users. Python is an excellent programming language for implementing machine learning and data mining algorithms, because it has a clear syntax; in Python it's very easy to manipulate text; Python uses a large number of people and organizations around the world, so it develops and is well documented; it is cross-platform and you can use it free. [7]

The analysis of the conversion series for compliance with ARMA models was performed. Figure 3 shows the typical values of the Akaike information criterion (AIC), which is used exclusively for the selection of several statistical models for one set of data [3]. The values of the criterion indicate that the use of ARMA models in this case is not appropriate.

| | Candidate | AIC |
|---|---|---|
| 1 | MAProcess[0] | 1411.31 |
| 2 | MAProcess[1] | 1412.81 |
| 3 | ARProcess[1] | 1412.85 |
| 4 | ARMAProcess[1, 1] | 1414.7 |

Fig.3. Values of the Akaike information criterion

To carry out the forecasting, time series were divided into two parts, where the first one was used to train the model, and the second one was applied to assess its plausibility. The models were trained on the $S$ last values of time series.

Checking the models for forecasting $m$ values was carried out in the following way: take the window of last $S$ values from the first part series and will do the forecast one value ahead; then will move window one value forward, including the forecast of new value in the window, and will again do the forecast, and so $m$ times.

Fig. 4 presents the results of the forecasts of each model for 7 values ahead or $S = 20$ and $m = 1$. The solid line shows the actual values. The upper part (a) shows the values obtained by the method of exponential smoothing, the average part (b) presents ones based on decision tree, in the bottom (c) you can see the values obtained with the help of the LSTM neural network.

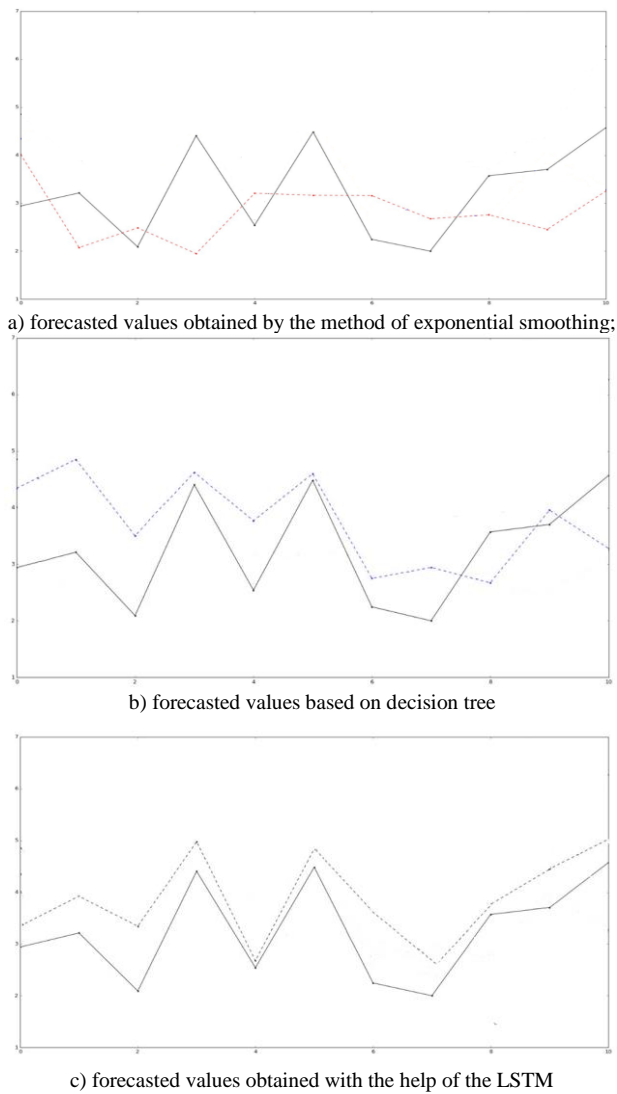

a) forecasted values obtained by the method of exponential smoothing;

b) forecasted values based on decision tree

c) forecasted values obtained with the help of the LSTM

Fig. 4. – Forecasts of each model

The predicted values for $S = 20$ and $m = 1$ (this choice of parameters is determined by the requirements of the online store) were computed for 100 values of the daily data conversion rate and corresponding values of clicks and sales number and other data. The results of calculations typical for most series are given in Table 1.

TABLE I. FORECAST ERRORS

|      | ES    | DT   | LSTM |
|------|-------|------|------|
| MAD  | 7.65  | 6.08 | 1.2  |
| MA   | -1.99 | 1.32 | 0.91 |
| MSE  | 64    | 53   | 22   |
| MAPE | 0.49  | 0.38 | 0.07 |

As a result of the analysis of the forecasts of different values of *S* and *m*, the following was established. The method of exponential smoothing, in spite of its simplicity and non-exactingness in the amount of data, has in most cases comparatively small prediction errors. But at the same time, with the use of this method, some predicted values are significantly removed from real ones. The decision tree method has proved to be inconvenient in the choice of parameters and has errors comparable with errors of exponential smoothing, but without strongly remote forecast values. The LSTM neural network, which has a more complex structure and needs to be preliminarily trained on a rather large time series, has shown good results, as well as in the overall forecast error, and in the remoteness of forecasts from real time series values.

## Conclusion

The results of a study of methods for predicting weakly correlated time series typical of e-commerce conversion series have shown that exponential smoothing is the simplest, fastest and most convenient to set up predictive method, but in the cases of complex or long-term dependencies, it does not apply. The decision tree method is fast in learning, not difficult to understand, but inconvenient in the choice of parameters and does not work well when learning on data that have many characteristics. The LSTM neural network is a cumbersome, long learning, requires a lot of parameters that need to be selected, but has a very good performance in forecasting and order of magnitude smaller errors.